# Deep ensembles in bioimage segmentation

Loris Nanni, Daniela Cuza, Alessandra Lumini, Andrea Loreggia and Sheryl Brahnam

**Abstract**— Semantic segmentation consists in classifying each pixel of an image by assigning it to a specific label chosen from a set of all the available ones. During the last few years, a lot of attention shifted to this kind of task. Many computer vision researchers tried to apply autoencoder structures to develop models that can learn the semantics of the image as well as a low-level representation of it. In an autoencoder architecture, given an input, an encoder computes a low dimensional representation of the input that is then used by a decoder to reconstruct the original data. In this work, we propose an ensemble of convolutional neural networks (CNNs). In ensemble methods, many different models are trained and then used for classification, the ensemble aggregates the outputs of the single classifiers. The approach leverages on differences of various classifiers to improve the performance of the whole system. Diversity among the single classifiers is enforced by using different loss functions. In particular, we present a new loss function that results from the combination of Dice and Structural Similarity Index. The proposed ensemble is implemented by combining different backbone networks using the DeepLabV3+ and HarDNet environment. The proposal is evaluated through an extensive empirical evaluation on two real-world scenarios: polyp and skin segmentation. All the code is available online at https://github.com/LorisNanni.

—————— ◆ ——————

## 1 INTRODUCTION

IN the computer vision domain, the task of semantic segmentation aims at classifying each pixel within an image. The purpose is to identify objects that are present in an image and their boundaries. The technique is used in many different contexts, from autonomous vehicles [1] to medical diagnosis [2]. The development of machine learning techniques with higher performances has attracted the attention of many researchers who are interested to see whether deep learning models can be used successfully also for these tasks. One of the first models employed for semantic segmentation leveraged on autoencoder architecture. In an autoencoder architecture, the input is processed by a module, which is called encoder, to produce a low dimensional representation of the input. This is then used to feed a decoder which is trained to reconstruct the original information. U-Net [3] was one of the first autoencoders developed for semantic segmentation. Unfortunately, it was unable to correctly identify the boundaries of objects. The issue was solved by mitigating the co-adaptation phenomenon by adding skip connections in the decoder module. Due to their success in the area of computer vision, autoencoders are employed in many tasks [4]–[6].

In this work, we propose a segmentation approach that is based on ensemble methods. Ensembles are groups of classifiers whose outputs are aggregated to form the prediction of the model. Roughly speaking, each classifier can be seen as a voter in an election which expresses its preference on a set of possible alternatives [7]. All the preferences are aggregated by the system to compute the output of the ensemble. This approach is justified by the "non-free lunch theorem" for machine learning: a single model that works well on all the instances of the same problem does not exist [8]. With this in mind, we developed ensembles of DeepLabV3+ for the purpose of image segmentation. In our approach, the sets of objects identified and segmented by the different classifiers may not overlap. This observation can be seen as another evidence that in ensembles different models can come up with different solutions for the same instance of a problem and their outputs might be complementary. The idea is that different models might learn different features from the training data and thus their aggregation is proved to lead to better performance. The key point in the implementation of ensembles is to enforce some kind of diversity in the set of classifiers; this can be seen as the ability of the set of classifiers of learning different features from the input space and thus reducing the error rate on the same input.

We provide an empirical evaluation of our system in two different scenarios. We tested the method against colorectal cancer segmentation and skin detection.

Being able to detect colorectal cancer in an early stage is of main importance to improve the probability of good results. It has been noticed that this pathology is somehow correlated with the presence of polyps. Thus, the detection and removal of polyps may help in reducing the appearance of colorectal cancer. Unfortunately, polyps are not easy to detect. This is due to the fact that the boundaries of polyps are very similar to the surrounding mucosa and thus they are difficult to be identified even by experienced medical experts. Polyps can be classified using the following taxonomy: 1) adenomatous, 2) serrated, 3) hyperplastic, 4) tubulovillous adenoma, and 5) inflammatory. Ade-

- L.N., D.C. are with the Department of Information Engineering, University of Padua, viale Gradenigo 6, 35122 Padua, Italy; E-mail: loris.nanni@unipd.it.
- A.L. is with the Department of Computer Science and Engineering, University of Bologna, Via dell'università 50, 47521 Cesena, Italy. E-mail: Alessandra.lumini@unibo.it.
- A.L. is with the Department of Legal Studies, University of Bologna - Via Zamboni, 33 - 40126 Bologna, Italy. Email: andrea.loreggia@gmail.com
- S.B is with Missouri State University, 901 South National Avenue Springfield, MO 65804, USA. SBrahnam@MissouriState.edu

nomas and serrated polyp are the most dangerous and difficult to detect. Machine learning techniques may help in the identification of these abnormal masses. Traditional classifiers were applied to the task of image segmentation with the aim of recognizing polyps segmentation [2], [9]–[11]. Recently, the application of CNNs to this task proved its superiority in different situations: CNNs were compared with standard classifiers to describe their better performance [12], suggesting also their adoption for developing segmentators [13] that won the first and second awards in the 2017 and 2018 Gastrointestinal Image ANAlysis (GIANA) contests.

The other side of the coin is that these models required a huge amount of data to be trained and such big datasets were not available for colorectal cancer image segmentations till some years ago [2], [9]. Recently, Jha et al. [14] published a new polyp dataset called Kvasir-SEG. The dataset contains 1,000 polyp images manually annotated at the pixel level. The annotation was carried out by expert endoscopists at Oslo University Hospital. The dataset was used to train a segmentator based on ResNet and U-Net [14]. The reported results show some very promising results.

In 2017 a novel approach based on varying the attention to the information lead to the development of a new model that is called Transformer [15]. Initially, it was designed and developed for natural language processing tasks (such as translation, semantic comprehension of texts, or summary creation). The astonishing performance of these models and the fact that their infrastructure seems to replicate the vision process in the human brain made these models appealing also for computer vision tasks. The structure of a Transformer is based on the idea of autoencoders with a self-attention mechanism that allows focusing only on some part of the input with high details while still considering at the same time the remaining parts of the input with low resolution. Due to the size of the model, the training phase is usually made in two steps: during the first step the model is trained using a large dataset (this is done to set the weights of the networks in order to generalize to a bigger solution space); in the second step, a fine-tuning of the model is performed using a smaller dataset (this is done to better fit the model to a specific domain) [16]. The complexity of the attention operator (which is quadratic) requires applying some reduction of the input size. In the computer vision domain, this is done by splitting the image into patches [17], after that, a linear transformation and position embeddings are applied. The resulting feature vectors are used as input for a transformer encoder. TransFuse [18] combines the ability of CNNs kernels of capturing local information with the ability of Transformers of representing information at a higher level. Another approach that tries to grasp the information at the local, as well as the global level, is UACANet [19]. It employs U-Net combined with a parallel axial attention encoder and decoder.

Skin detection is one of the most important tasks faced by researchers who want to implement models able to recognize human activity. From video monitoring, to face detection, from hand gesture recognition to content-based detection, practitioners start investigating the use of deep learning techniques applied to skin detection. This is due to the important improvements that these techniques have in many different domains. Nevertheless, these approaches have to deal with many challenges. For instance, robust detection of hand gestures in real-world environments is hindered by issues connected to the background clutter. To overcome such difficulties, Roy et al. [20] propose to enhance the output of the hand detector by using a convolutional neural network (CNN) based on skin detection technique to significantly reduce occurrences of false positives. To deal with skin segmentation in difficult situations and to reduce the computational cost, Arsalan et al. [21] propose a residual skip connection-based deep convolutional neural network (OR-Skip-Net) that transfers information from the initial layer to the last layer of the network. Skin detection is once again employed in Shahriar et al. [22] to develop an automatic American Sign Language (ASL) fingerspelling translator based on a convolutional neural network. Lumini and Nanni [23] report an extensive empirical analysis on different datasets to investigate how the available technology performs and propose a fair comparison among approaches in order to reduce discrimination.

**Contribution.** In this work, we design and develop ensemble classifiers to be used for semantic segmentation. We show that diversity in the ensemble can be enforced varying the loss functions. In particular, we propose a new loss function resulting from the combination of Dice function and a structural similarity index. We report an extensive empirical analysis showing the good performance of our approach. We also compare our ensemble model with state-of-the-art methods, considering also the recently proposed HarDNet-MSEG [24] and transformer-based approaches.

## 2 METHODS

In the following section, we give a brief description of the adopted methods and a mathematical definition of the several loss functions and metrics employed in this work.

### 2.1 Deep Learning for Semantic Image Segmentation

In this section, we shall focus on the application of deep learning methods on the task of semantic segmentation. As previously described, semantic segmentation consists in assigning a class to each object in an image. The classification process is done at the pixel level. Fully Convolutional Networks (FCNs) were initially adopted to solve the problem of semantic segmentation. In FCNs, the last fully connected layer is replaced with a fully convolutional layer that allows the network to work at the pixel level [6]. The introduction of encoder-decoder unit into FCNs allows designing, developing, and training deconvolutional net-works. Some examples of autoencoders developed for semantic segmentation are: a) U-Net [25], which employs autoencoder to downsample the image but at the same time increases the number of features that represent the input and increase the resolution for computing the segmentation; b) SegNet [4], adopt VGG [26] for the encoder module. Instead of feeding the decoder module with the output of the encoder, each SegNet decoder layer uses max pool indices of the corresponding encoder layer. This results in less memory consumption and better output segmentation; c) DeepLab [27] is a family of models designed by Google that gives good performance in the semantic segmentation field. The model upsamples the output of the last convolution layer through an atrous convolution process. This consists in applying a dilation rate to increase the window of filters, but at the same time maintaining a stable computational effort. This family of segmentators [27]–[30] reaches better performances due to:

- the dilated convolution that maintains a high-resolution despite the pooling and stride effects;
- Atrous Spatial Pyramid Pooling that allows to retrieve information at different scales;
- a combination of CNNs and probabilistic graphical models that allows for a better localization of object boundaries.

Two are the novelties introduced in a new version of the family, called DeepLabV3: 1) a combination of cascade and parallel modules for convolutional dilation; 2) the addition of batch normalization and 1x1 convolution in Atrous Spatial Pyramid Pooling.
Another extension of the Google family is DeepLabV3+ [30] (which is the framework adopted in this work). It includes 1) a decoder with point-wise convolutions that operates on the same channel but at different locations and 2) a depth-wise convolutions that operates at the same location but on different channels.
Many other deep learning models have been applied to the task of image segmentation. A non-exhaustive list comprehends recurrent neural networks, attention models, generative approaches. We point the interested reader to a recent survey [31].
Model architecture is just one of the possible choices available to influence the design of a framework. In this work, we explore ResNet18 and ResNet50 [32], two well-known CNN architectures that learn a residual function with reference to the block input (for a comprehensive list of CNN structures, we point the interested reader to [33]).
Transfer learning in the form of pretrained encoders is another possible choice that can impact the design of a model. The application of different loss functions instead would affect the training phase of the model.

In the image segmentation field, one of the most adopted loss functions is the pixel-wise cross-entropy loss. With this loss function, the classification task works at the pixel level by confronting the predicted label for a pixel with the real one. The downside is that this function does not work properly when the dataset is unbalanced in favor of one or more classes. Counterweights can be applied to reduce the problem and to cope with the imbalance in the dataset [6]. Measuring the overlap among segmented images is the goal of Dice loss [34], which is derived from the Sørensen-Dice similarity coefficient. This is a common loss function adopted in literature and the one employed in this work. The function range is in [0, 1], where 1 means a perfect overlap between two masks. The reader is referred to [34] for an overview of other popular loss functions for image segmentation. In Section 2.2, the loss functions used in our experiments will be presented, including a novel loss for image segmentation named SSimDice.

## 2.2 Loss functions

Dice Loss comes from Sørensen-Dice coefficient, a metric widely used to estimate the performance of semantic segmentation models. The Sørensen-Dice coefficient shows how similar two images are to each other in a range [0, 1]. In order to apply dice loss to multiclass problems, the Generalized Dice Loss was proposed [35]. The formula of the Generalized Dice Loss between the predictions $Y$ and the training targets $T$ is:

$$L_{GD}(Y,T) = 1 - \frac{2 * \sum_{k=1}^{K} w_k * \sum_{m=1}^{M} Y_{km} * T_{km}}{\sum_{k=1}^{K} w_k * \sum_{m=1}^{M} (Y_{km}^2 + T_{km}^2)} \quad (1)$$

$$w_k = \frac{1}{\left(\sum_{m=1}^{M} T_{km}\right)^2} \quad (2)$$

where $K$ is the number of classes, $M$ is the number of pixels and the weighting factor $w_k$ is introduced to help the network focuses on a small region. Indeed, it is inversely proportional to the frequency of the labels of a given class $k$.
A recurring problem in image segmentation is the predominance of one class over another. In the interest of dealing with this issue, Tversky Loss was introduced [36]. Tversky Loss derives from Tversky Index, which can be seen as an extension of the dice similarity coefficient. Tversky Index has two weighting factors, $\alpha$ and $\beta$, in order to manage trade-off between false positives and false negatives. When $\alpha = \beta = 0.5$ the Tversky Index degenerates into Dice Similarity coefficient. The Tversky Index between the predictions $Y$ and the ground truth $T$ for a given class $k$ is:

$$TI_k(Y,T) = \frac{\sum_{m=1}^{M} Y_{pm} T_{pm}}{\sum_{m=1}^{M} Y_{pm} T_{pm} + \alpha \sum_{m=1}^{M} Y_{pm} T_{nm} + \beta \sum_{m=1}^{M} Y_{nm} T_{pm}} \quad (3)$$

where $p$ indicates the positive class, $n$ the negative class, $M$ the total number of pixels. The formula of the Tversky Loss is:

$$L_T(Y,T) = \sum_{k=1}^{K} (1 - TI_k(Y,T)) \quad (4)$$

where $K$ is the number of classes.
In this work, we have set the $\alpha = 0.3$ and $\beta = 0.7$, i.e. we have given more emphasis to false negatives.
One of the most popular distribution-based loss functions is cross-entropy (CE). CE aims to minimize the difference between two probability distributions, and it has no bias between large and small regions. Several variants of the cross-entropy loss have been proposed in literature, such as Binary Cross-Entropy and Focal loss [37]. Binary Cross-Entropy is simply the application of the CE to problems with two classes. The main purpose of the Focal loss is to focus the model on hard examples and down-weight well-classified examples by adding a modulating factor $\gamma > 0$. Focal Loss ($L_F$) works well when there is a high imbalance between foreground and background classes. Other loss functions that uses $\gamma$ factor in order to learn hard-examples are Focal Tversky Loss [38] and Exponential Logarithmic Loss [39].

$$L_{FT}(Y,T) = L_T(Y,T)^{\frac{1}{\gamma}} \quad (5)$$

Focal Tversky Loss achieves a good trade-off between precision and recall by using Tversky Index.

Influenced by the Focal Tversky Loss, we applied the modulating factor $\gamma$ also to Generalized Dice Loss. Focal Generalized Dice Loss down-weight common examples and concentrates on small ROIs. In our experiments we set $\gamma = 4/3$.

$$L_{FGD}(Y,T) = L_{GD}(Y,T)^{\frac{1}{\gamma}} \quad (6)$$

Log-Cosh Dice Loss derives from Dice Loss and Log-Cosh function, used in regression problems for smoothing the curve. In fact, $\log(\cosh(x))$ is approximately equal to $x^2/2$ for small $x$ and to $|x| - \log(2)$ for large $x$. Log-Cosh Generalized Dice Loss is given by:

$$L_{lcGD}(Y,T) = \log(\cosh(L_{GD}(Y,T))) \quad (7)$$

Inspired by Log-Cosh Dice Loss, in our experiments we try to smooth other loss function curves. In particular, we propose Log-Cosh Binary Cross Entropy Loss, Log-Cosh Tversky Loss and Log-Cosh Focal Tversky Loss, which can be defined as:

$$L_{lcFT}(Y,T) = \log(\cosh(L_{FT}(Y,T))) \quad (8)$$

These are variants of Binary Cross-Entropy Loss, Tversky Loss and Focal Tversky Loss respectively. The difference is the addition of the term Log-Cosh.

Recently, the Neighbor Loss [40] has been introduced. It can be seen as a weighted cross-entropy, where all pixels have a different weight according to its 8 neighbors, in order to consider the spatial correlation of neighborhood. The weight of a pixel depends on the number of neighbors that have a different prediction from the center one. Like Focal Loss, Neighbor Loss tries to deal with hard samples. It uses a threshold $t$ and a binary indicator function $1\{\cdot\}$, in order to drop easily classified pixels. In our tests it obtains low performance, and it is not considered in the proposed ensemble.

SSIM Loss [41], that is used to estimate the quality of an image, derives from Structural similarity (SSIM) index [42]. It is defined as:

$$SSim(x,y) = \frac{(2\mu_x\mu_y + C_1)(2\sigma_{xy} + C_2)}{(\mu_x^2 + \mu_y^2 + C_1)(\sigma_x^2 + \sigma_y^2 + C_2)} \quad (9)$$

where $\mu_x$, $\mu_y$, $\sigma_x$, $\sigma_y$, and $\sigma_{xy}$, are respectively the local means, the standard deviations, and the cross-covariance for images x, y, while $C_1$, $C_2$ are regularization constants. The SSIM Loss between one image $Y$ and the corresponding ground truth $T$ is given by:

$$L_S(Y,T) = 1 - SSim(Y,T) \quad (10)$$

In applications with unbalanced data, such as the early detection of a cancer, the risk is to achieve high precision but low recall. Generalized Dice Loss uses a recurrent technique to decrease the effects of class imbalance: it adds a weight $w_k$, which is the inverse of label frequency. One limitation of the Dice Loss is the fact that it is a harmonic mean of FPs and FNs. To guarantee that no lesion is missed, we need flexibility in balancing FPs and FNs, in particular doctors manage to weight FNs higher than FPs. In order to focus the model on hard examples and to incorporate benefits of both Focal Generalized Dice Loss and Focal Tversky Loss, we combined them:

$$Comb_1(Y,T) = L_{FGD}(Y,T) + L_{FT}(Y,T) \quad (11)$$

Another idea to down-weight easy examples is to mix Log-Cosh Dice Loss, Focal Generalized Dice Loss and Log-Cosh Focal Tversky Loss. In this case we also control the non-convex nature of the curve by using Log-Cosh approach:

$$Comb_2(Y,T) = L_{lcGD}(Y,T) + L_{FGD}(Y,T) + L_{lcFT}(Y,T) \quad (12)$$

Finally we propose a combination of the SSIM Loss and the Generalized Dice Loss:

$$Comb_3(Y,T) = L_S(Y,T) + L_{GD}(Y,T) \quad (13)$$

### 2.3 Stochastic Activation Selection

Recently, Lumini and Nanni [43] introduced a new process to generate different networks from a specific one. This is called stochastic activation selection. Basically, given a neural network the process generates a new one that has the same architecture but differs from the original for the adopted activation function. This is chosen at random from a set of available ones.

Reiterating the process, we can generate as many networks as we need, all with the same structure but different activation functions. After the generation of the networks, they are trained on the same set of samples and their prediction combined using the sum rule. In the empirical evaluation, we use DeepLabV3+ [30] neural network architecture, and we draw at random the activation function from ReLU [44], Leaky ReLU [45], ELU [46], PReLU [47], S-Shaped ReLU (SReLU) [48] and many other. For more details about the whole set of used activation functions, we point out the reader to [49].

## 3 EXPERIMENTAL RESULTS

In this section, we report the results of the empirical evaluation of the proposed ensemble methods as well as the comparison with state-of-the-arts models.

### 3.1 Datasets, testing protocol and metrics - Polyp segmentation

We run the experiments on five polyp datasets. These datasets are well-known cases widely used in literature for evaluation purposes. These datasets are:
- Kvasir-SEG [14] (**Kvasir**): it includes 1000 polyp images acquired by a high-resolution electromagnetic imaging system: 900 are used for training, the remaining tor testing.

- CVC-ColonDB [50] (**ColDB**): it consists of 380 images (574×500) representing 15 different polyps;
- EndoScene-CVC300 (**CVC-T**): it is the test set of a large dataset [51];
- ETIS-Larib Polyp DB [52] (**ETIS**): it consists of frames extracted from colonoscopy videos, annotated by expert video endoscopists and includes 196 high-resolution images (1225×966);
- CVCClinic DB [53] (**ClinDB**): it consists of frames extracted from colonoscopy videos, annotated by expert video endoscopists and includes 612 images (384×288).

The testing protocol includes a training set of 1450 images (900 images from Kvasir and 550 images from ClinDB) and a 5 testing sets including the remaining images from the above cited datasets (100 images from Kvasir, 62 images from ClinDB, 380 from ColonDB, 196 ETIS and 60 from CVC-T). All the images, already divided in training and testing set are available in github[1] [24].

In the empirical evaluation, it might be necessary to resize the input to fit the input size of the adopted model. The predicted masks are always resized back to the original dimensions. In this work, we report only the performance evaluation made on the original size of the masks. We do not consider approaches that evaluate the model on different sizes of images.

As already done in literature, we adopt the following metrics to evaluate the performance of the model:
1. Dice: twice the overlap area of the predicted and ground-truth masks divided by the total number of pixels (it coincides to the F1score for binary masks, i.e. a weighted average of precision and recall).
2. Intersection over Union (IoU): the area shared between the predicted mask A and the ground truth map B, divided by the area of the union between the two maps;

A mathematical definition for each of the aforementioned metrics is given:

$$F1score = Dice = \frac{|A \cap B|}{|A| + |B|} = \frac{2 \cdot TP}{2 \cdot TP + FP + FN} \quad (1)$$

$$IoU = \frac{|A \cap B|}{|A \cup B|} = \frac{TP}{TP + FP + FN} \quad (2)$$

where, considering a bi-class problem (foreground/background), TP, TN, FP, FN refer to the true positives, true negatives, false positives, and false negatives, respectively.

### 3.2 Experiments on Polyp segmentation

The first experiment is aimed at comparing the different loss functions, presented in Section 2.2, that are used in the ensembles. According to the findings in [54], ResNet18 and ResNet50 have been selected as backbone decoders and the size of input images is fixed to 352. The results from stand-alone DeepLabV3+ coupled ResNet18 (*RN18*) and ResNet50 (*RN50*) are considered as baselines (i.e. CNNs based on standard ReLu layers).

In Tables 1 and 2, the performance of a stochastic network with backbone Resnet18 (Table 1) and ResNet50 (Table 2) are reported as a function of the loss function used to train the net.

TABLE 1
EXPERIMENTS USING RESNET18 AS BACKBONE (DICE)

|  | Kvasir | ClinDB | ColDB | ETIS | CVC-T | Avg |
|---|---|---|---|---|---|---|
| *RN18* | 0.888 | 0.914 | 0.74 | 0.631 | 0.844 | 0.803 |
| $L_{GD}$ | 0.89 | 0.903 | 0.733 | **0.661** | 0.845 | **0.806** |
| $L_F$ | 0.87 | 0.847 | 0.669 | 0.58 | 0.823 | 0.758 |
| $L_T$ | 0.879 | 0.899 | 0.739 | 0.633 | 0.854 | 0.801 |
| $L_{FT}$ | 0.88 | 0.903 | **0.738** | 0.602 | **0.866** | 0.798 |
| $L_{FGD}$ | 0.888 | 0.892 | 0.735 | 0.629 | 0.853 | 0.799 |
| $Comb_1$ | 0.897 | 0.901 | 0.719 | 0.616 | 0.85 | 0.797 |
| $Comb_2$ | 0.875 | 0.885 | 0.728 | 0.623 | 0.84 | 0.790 |
| $Comb_3$ | **0.893** | **0.917** | 0.737 | 0.636 | 0.848 | **0.806** |

TABLE 2
EXPERIMENTS USING RESNET50 AS BACKBONE (DICE)

|  | Kvasir | ClinDB | ColDB | ETIS | CVC-T | Avg |
|---|---|---|---|---|---|---|
| *RN50* | 0.895 | 0.898 | 0.716 | 0.591 | **0.892** | 0.798 |
| $L_{GD}$ | 0.892 | 0.892 | 0.716 | 0.632 | 0.878 | 0.802 |
| $L_F$ | 0.869 | 0.832 | 0.602 | 0.449 | 0.825 | 0.715 |
| $L_T$ | 0.897 | 0.885 | 0.721 | 0.594 | 0.865 | 0.792 |
| $L_{FT}$ | 0.887 | 0.897 | 0.708 | **0.673** | 0.868 | **0.807** |
| $L_{FGD}$ | 0.886 | 0.9 | 0.686 | 0.626 | **0.882** | 0.796 |
| $Comb_1$ | 0.9 | 0.894 | 0.711 | 0.606 | 0.822 | 0.787 |
| $Comb_2$ | 0.878 | **0.904** | **0.726** | 0.629 | 0.84 | 0.795 |
| $Comb_3$ | **0.894** | 0.903 | 0.719 | 0.614 | 0.868 | 0.800 |

From the results in Tables 1 and 2 we notice that there is not a clear winner among the loss functions. The diversity among the resulting networks and the possibility of creating an ensemble based on the fusion of these network is explored in the second experiment.

The second experiment is aimed at evaluating ensembles. The only stand-alone approach is *RN101,* a DeepLabV3+ with ResNet101 backbone[2] pretrained on VOC (generalized loss is used as loss function); this network has an input size (513) larger than the previous ones. In Table 3, the performance of eight ensembles of 10 classifiers are included: two are obtained by models based on standard ReLu activation function (i.e. *ERN18(10)*, *ERN50(10)*), two are the fusion of stochastic model obtained as described in section 2.3 (i.e. *ESto18(10)*, *ESto50(10)*). The method named *ERN101*(10) is the fusion of 10 *RN101* (due to computation

---

[1] https://github.com/james128333/HarDNet-MSEG
[2] https://github.com/matlab-deep-learning/pretrained-deeplabv3plus notice that this network is trained using the parameters suggested in the github page, to avoid any overfitting we have not modified them

issues the stochastic version of this network has not been included).

Moreover, the following ensembles obtained by varying the loss functions are reported:

- *ELoss18*(10) is an ensemble of 10 stochastic networks with backbone Resnet18 trained using different loss functions: 2×$L_{GD}$+2×$L_T$+2× $Comb_1$+2× $Comb_2$+2× $Comb_3$
- *ELoss50*(10) is an ensemble of 10 stochastic networks with backbone Resnet50 trained using different loss functions: 2×$L_{GD}$+2×$L_T$+2× $Comb_1$+2× $Comb_2$+2× $Comb_3$
- *ELoss101*(10) is an ensemble of 10 pretrained Resnet101 trained using different loss functions: 2×$L_{GD}$+2×$L_T$+2× $Comb_1$+2× $Comb_2$+2× $Comb_3$

Notice that all the ensembles use the simple sum rule to aggregate the predictions of internal networks.

The training options (when ResNet18/50 are used as backbone) are the following: SGD optimizer, 20 epochs, learning rate 10e-2, learning drop factor of 0.2 every 5 epochs. Data augmentation is performed as in [54], i.e. horizontal and vertical flip, 90° rotation.

TABLE 3
EXPERIMENTS USING ENSEMBLES (DICE)

|  | Kvasir | ClinDB | ColDB | ETIS | CVC-T | Avg |
|---|---|---|---|---|---|---|
| *RN101* | 0.881 | 0.889 | 0.756 | 0.657 | 0.855 | 0.808 |
| *ERN18*(10) | 0.898 | 0.924 | 0.768 | 0.660 | 0.857 | 0.821 |
| *ERN50*(10) | 0.904 | 0.911 | 0.712 | 0.615 | **0.891** | 0.807 |
| *ESto18*(10) | 0.896 | 0.921 | 0.763 | 0.642 | 0.851 | 0.815 |
| *ESto50*(10) | 0.909 | 0.906 | 0.729 | 0.617 | **0.891** | 0.810 |
| *ERN101*(10) | 0.899 | 0.916 | 0.783 | 0.712 | 0.862 | 0.834 |
| *ELoss18*(10) | 0.905 | 0.921 | 0.763 | 0.671 | 0.860 | 0.824 |
| *ELoss50*(10) | 0.912 | 0.909 | 0.742 | 0.632 | 0.882 | 0.815 |
| *ELoss101*(10) | 0.912 | 0.927 | 0.763 | **0.719** | **0.891** | **0.843** |

From the results in Table 3, it is clear that ensembles are useful to improve the performance of stand-alone approaches. In this segmentation problem there is only a slight difference between ensembles obtained by models based on standard ReLu activation function and those obtained by the fusion of stochastic models. Finally, ResNet18 seems to work slightly better than ResNet50 in ensembles, while ResNet101 overcomes both. The use of a network pretrained on a segmentation problem allows a further improvement, but at the expense of higher computational costs: *ERN101* and *ELoss101* have a larger input size, and they are based on a deeper network with respect to ResNet18 and ResNet50.

Comparing the results of the last three ensembles (*ELoss\**) with those where networks use all the same loss function, it is evident that the variation of the loss is useful to create diversity among networks, resulting in a slightly better performance.

Finally, for the sake of comparison, in Tables 4 and 5 we report the performance of several state-of-the-art-approaches evaluated using both Dice and IoU as performance indicators. Among the dozens of methods in the literature evaluated using this protocol, we report only the best approaches to be compared with the following methods:

- *HN_SGD* and *HD_Adam* are our experiments using HardNet [24] trained by SGD and Adam optimizers (using the code shared by the authors);
- *HN_A&S* is the fusion of *HN_Adam* + *HN_SGD*;
- *Res18&50* is the fusion of *ELoss18(10)* + *ELoss50(10)*;
- *Res18&50&101* is the fusion of *ELoss18(10)* + *ELoss50(10)* + *ELoss101(10)*;
- *HN&101* is the fusion of 10×*HN_A&S* + *ELoss101(10)*, the weight of *HN_A&S* is 10 since *ELoss101(10)* is the sum rule among 10 networks;
- *EnsAll* is the fusion of *ELoss18(10)* + *ELoss50(10)* + *ELoss101(10)* + 30×*HN_A&S*, the weight of *HN_A&S* is 30 since *ELoss18(10)* + *ELoss50(10)* + *ELoss101(10)* is the sum rule among 30 networks.

TABLE 4
STATE-OF-THE-ART COMPARISONS (DICE)

|  | Kvasir | ClinDB | ColDB | ETIS | CVC-T | Avg |
|---|---|---|---|---|---|---|
| *HarDNet_SGD* | 0.908 | 0.911 | 0.752 | 0.639 | 0.868 | 0.816 |
| *HarDNet_Adam* | 0.906 | 0.924 | 0.751 | 0.716 | 0.903 | 0.840 |
| *HN_A&S* | 0.915 | 0.926 | 0.776 | 0.743 | 0.894 | 0.851 |
| *Res18&50* | 0.911 | 0.926 | 0.765 | 0.668 | 0.876 | 0.829 |
| *Res18&50&101* | 0.915 | 0.930 | 0.771 | 0.683 | 0.884 | 0.837 |
| *EnsAll* | 0.919 | 0.931 | 0.772 | 0.725 | 0.899 | 0.849 |
| *HN&101* | 0.917 | 0.931 | 0.769 | 0.740 | 0.901 | 0.852 |
| *HarDNet* [24] | 0.912 | 0.932 | 0.731 | 0.677 | 0.887 | 0.828 |
| *PraNet* [24] | 0.898 | 0.899 | 0.709 | 0.628 | 0.871 | 0.801 |
| *SETR* [55] | 0.911 | 0.934 | 0.773 | 0.726 | 0.889 | 0.847 |
| *TransUnet* [56] | 0.913 | 0.935 | **0.781** | 0.731 | 0.893 | 0.851 |
| *TransFuse* [18] | **0.920** | **0.942** | **0.781** | 0.737 | 0.894 | **0.855** |
| *UACANet* [19] | 0.912 | 0.926 | 0.751 | **0.751** | **0.910** | 0.850 |

TABLE 5
STATE-OF-THE-ART COMPARISONS (IOU)

|  | Kvasir | ClinDB | ColDB | ETIS | CVC-T | Avg |
|---|---|---|---|---|---|---|
| *HarDNet_SGD* | 0.857 | 0.864 | 0.677 | 0.562 | 0.799 | 0.752 |
| *HarDNet_Adam* | 0.854 | 0.875 | 0.678 | 0.625 | 0.831 | 0.772 |
| *HN_A&S* | 0.866 | 0.880 | 0.700 | 0.652 | 0.823 | 0.784 |
| *Res18&50* | 0.856 | 0.880 | 0.691 | 0.605 | 0.805 | 0.767 |
| *Res18&50&101* | 0.862 | 0.885 | 0.700 | 0.623 | 0.815 | 0.777 |
| *EnsAll* | **0.872** | 0.886 | 0.701 | 0.650 | 0.829 | 0.788 |
| *HN&101* | 0.871 | 0.886 | 0.697 | 0.663 | 0.831 | 0.790 |
| *HarDNet* [24] | 0.857 | 0.882 | 0.66 | 0.613 | 0.821 | 0.767 |
| *PraNet* [24] | 0.84 | 0.849 | 0.64 | 0.567 | 0.797 | 0.739 |
| *SETR* [55] | 0.854 | 0.885 | 0.69 | 0.646 | 0.814 | 0.778 |
| *TransUnet* [56] | 0.857 | 0.887 | 0.699 | 0.660 | 0.824 | 0.785 |
| *TransFuse* [18] | 0.870 | **0.897** | **0.706** | 0.663 | 0.826 | **0.792** |
| *UACANet* [19] | 0.859 | 0.880 | 0.678 | **0.678** | **0.849** | 0.789 |

Among the ensembles proposed in this work, *HN&101* gains the best results, with performance better than methods based on transformers (i.e. TransFuse and TransUnet) in at least 2 out of 5 datasets. HN&101 boosts performance of convolutional neural networks up to performance of transformer-based approaches.
Notice that HarDNet [24] is the state of the art among CNN based approaches (see the comparisons reported in that paper), it is interesting to notice that:
- HN_A&S outperforms the stand-alone HarDNet;
- HN&101 outperforms HN_A&S.

The same conclusions obtained in the polyp datasets are also obtained in the skin datasets, see Table 10 showing the robustness of the proposed ensemble.

### 3.3 Experiments on skin segmentation

Skin detection is a binary-class segmentation problem that segments images into "skin" and "nonskin" regions. A wide used testing protocol for skin segmentation is the one proposed in [22] where methods are trained on a small training set made up of the first 2000 labeled images, from the well-known ECU dataset [57], and tested on 10 datasets (Table 6) for skin segmentation.

The performance indicator used in the literature is F1-score, i.e. Dice, calculated at pixel-level (and not at image-level) to be independent from the image size of the different datasets.

TABLE 6
DATASETS FOR SKIN SEGMENTATION

| ShortName | Name | #Samples | Ref. |
|---|---|---|---|
| Prat | Pratheepan | 78 | [58] |
| MCG | MCG-skin | 1000 | [59] |
| UC | UChile DB-skin | 103 | [60] |
| CMQ | Compaq | 4675 | [61] |
| SFA | SFA | 1118 | [62] |
| HGR | Hand Gesture Recognition | 1558 | [63] |
| Sch | Schmugge dataset | 845 | [64] |
| VMD | Human activity recognition | 285 | [65] |
| ECU | ECU Face and Skin Detection | 4000 | [57] |
| VT | VT-AAST | 66 | [66] |

Tables 7 reports the results obtained using ResNet18 varying the activation function, and Table 8 the results obtained with a ResNet50 varying the activation function, while the evaluation of the considered ensembles in the second polyp experiments is reported in Table 9.

To reduce the computation time, in the tests for skin segmentation, we did not use stochastic ensemble but we employed only networks based on ReLu layers.

TABLE 7
EXPERIMENTS ON SKIN BASED ON RESNET18 (F1SCORE)

|  | Prat | MCG | UC | CMQ | SFA | HGR | Sch | VMD | ECU | VT | Avg |
|---|---|---|---|---|---|---|---|---|---|---|---|
| $L_{GD}$ | 0.91 | 0.882 | 0.87 | 0.834 | 0.947 | 0.961 | 0.76 | **0.737** | 0.946 | **0.807** | 0.865 |
| $L_F$ | 0.896 | 0.874 | 0.882 | 0.827 | 0.946 | 0.955 | 0.76 | 0.709 | 0.94 | 0.756 | 0.855 |
| $L_T$ | 0.9 | **0.888** | 0.886 | 0.83 | 0.949 | 0.963 | 0.761 | 0.701 | 0.942 | 0.802 | 0.862 |
| $L_{FT}$ | 0.904 | 0.886 | **0.899** | 0.829 | 0.951 | 0.965 | **0.773** | 0.709 | 0.945 | 0.802 | 0.866 |
| $L_{FGD}$ | 0.908 | 0.88 | 0.872 | 0.834 | 0.948 | 0.964 | 0.765 | 0.705 | 0.944 | 0.795 | 0.862 |
| $Comb_1$ | 0.906 | 0.882 | 0.881 | 0.833 | 0.949 | 0.961 | 0.765 | 0.693 | 0.946 | 0.771 | 0.859 |
| $Comb_2$ | 0.911 | 0.884 | 0.882 | 0.835 | 0.95 | 0.962 | 0.77 | 0.721 | 0.946 | 0.789 | 0.865 |
| $Comb_3$ | **0.912** | 0.887 | 0.877 | **0.846** | **0.953** | **0.969** | **0.773** | 0.718 | **0.949** | 0.784 | **0.867** |

TABLE 8
EXPERIMENTS ON SKIN BASED ON RESNET50 (F1SCORE)

|  | Prat | MCG | UC | CMQ | SFA | HGR | Sch | VMD | ECU | VT | Avg |
|---|---|---|---|---|---|---|---|---|---|---|---|
| $L_{GD}$ | 0.909 | **0.885** | 0.9 | **0.838** | 0.95 | **0.966** | **0.779** | **0.741** | 0.946 | **0.794** | **0.871** |
| $L_F$ | 0.883 | 0.861 | 0.82 | 0.786 | 0.93 | 0.936 | 0.684 | 0.592 | 0.929 | 0.682 | 0.810 |
| $L_T$ | 0.907 | 0.884 | 0.88 | 0.828 | 0.946 | 0.96 | 0.747 | 0.694 | 0.945 | 0.745 | 0.854 |
| $L_{FT}$ | 0.911 | 0.883 | **0.903** | 0.829 | 0.948 | 0.963 | 0.756 | 0.705 | 0.945 | 0.741 | 0.858 |
| $L_{FGD}$ | 0.909 | 0.881 | 0.89 | 0.83 | 0.945 | 0.964 | 0.677 | 0.697 | 0.946 | 0.73 | 0.847 |
| $Comb_1$ | **0.913** | 0.881 | 0.897 | 0.83 | 0.944 | 0.963 | 0.766 | 0.703 | 0.946 | 0.744 | 0.859 |
| $Comb_2$ | 0.91 | 0.88 | 0.896 | 0.827 | 0.944 | 0.959 | 0.744 | 0.696 | 0.945 | 0.732 | 0.853 |
| $Comb_3$ | 0.911 | 0.88 | 0.884 | 0.83 | 0.947 | 0.959 | 0.729 | 0.671 | **0.947** | 0.729 | 0.849 |

As clearly shown by F1-scores reported in Tables 7-8, Comb$_3$ is the best approach when coupled with ResNet18. On the contrary, when ResNet50 is used as backbone, the best loss is the generalized Dice. This is especially true in the VMD and VT datasets where it strongly outperforms all the other tested loss.

TABLE 9
ENSEMBLE FOR SKIN SEGMENTATION (F1SCORE)

|  | Prat | MCG | UC | CMQ | SFA | HGR | Sch | VMD | ECU | VT | Avg |
|---|---|---|---|---|---|---|---|---|---|---|---|
| RN101 | 0.922 | 0.887 | **0.923** | 0.823 | 0.948 | 0.969 | 0.750 | 0.748 | 0.948 | 0.796 | 0.871 |
| ERN18 | 0.91 | 0.882 | 0.889 | 0.838 | 0.949 | 0.96 | 0.76 | 0.734 | 0.946 | 0.795 | 0.866 |
| ERN50 | 0.913 | 0.885 | 0.888 | **0.846** | 0.95 | 0.968 | **0.781** | 0.750 | 0.949 | 0.797 | 0.872 |
| ERN101 | 0.924 | 0.887 | 0.920 | 0.845 | 0.952 | **0.971** | 0.778 | **0.754** | 0.950 | 0.794 | 0.878 |
| ELoss18(10) | 0.915 | 0.889 | 0.894 | **0.846** | 0.953 | 0.969 | 0.776 | 0.735 | 0.950 | 0.797 | 0.872 |
| ELoss50(10) | 0.917 | 0.887 | 0.896 | 0.840 | 0.951 | 0.967 | 0.765 | 0.714 | 0.951 | 0.746 | 0.863 |
| ELoss101(10) | **0.926** | **0.892** | **0.923** | 0.844 | **0.956** | **0.971** | 0.777 | 0.751 | **0.953** | **0.807** | **0.880** |

From the results reported in Table 9 the following conclusions can be drawn:
- *ELoss18(10)* and *Eloss101(10)* outperforms, respectively, ERN18 and ERN101;
- *Eloss50(10)* has a lower performance than ERN50, mainly due to VMD and VT datasets;
- Similarly to experiments in the polyp domain, the best approach is *Eloss101(10)*.

Finally, in Table 10 the comparison with some state-of-the-art approaches is reported. The results for HarDNet and for all other methods have been calculated using the same parameters configuration used for the Polyp experiments, with the aim to avoid overfitting. Our ensembles are the same used for polyp segmentation.

TABLE 10
STATE-OF-THE-ART COMPARISONS (F1SCORE)

|  | Prat | MCG | UC | CMQ | SFA | HGR | Sch | VMD | ECU | VT | Avg |
|---|---|---|---|---|---|---|---|---|---|---|---|
| HarDNet_SGD | 0.903 | 0.880 | 0.903 | 0.838 | 0.947 | 0.964 | 0.793 | 0.744 | 0.941 | 0.810 | 0.872 |
| HarDNet_Adam | 0.913 | 0.880 | 0.900 | 0.809 | 0.951 | 0.967 | 0.792 | 0.717 | 0.945 | 0.799 | 0.867 |
| HN_A&S | 0.916 | 0.886 | 0.911 | 0.838 | 0.952 | 0.969 | **0.800** | 0.754 | 0.948 | 0.817 | 0.879 |
| Res18&50 | 0.920 | 0.890 | 0.896 | 0.846 | 0.952 | 0.969 | 0.775 | 0.728 | 0.954 | 0.767 | 0.870 |
| Res18&50&101 | 0.924 | **0.891** | 0.906 | **0.848** | **0.955** | **0.971** | 0.775 | 0.741 | **0.954** | 0.782 | 0.875 |
| EnsAll | 0.922 | 0.888 | 0.914 | 0.845 | 0.954 | **0.971** | 0.798 | 0.761 | 0.950 | 0.817 | 0.882 |
| HN&101 | **0.926** | 0.888 | **0.916** | 0.842 | **0.955** | **0.971** | 0.799 | **0.764** | 0.952 | **0.820** | **0.883** |
| HarDNet-352 [54] | 0.913 | 0.887 | 0.902 | 0.835 | 0.952 | 0.968 | 0.795 | 0.729 | 0.946 | 0.744 | 0.867 |
| SelEns [54] | 0.917 | 0.884 | 0.910 | 0.840 | 0.952 | 0.968 | 0.785 | 0.742 | 0.949 | 0.755 | 0.870 |
| FusAct3 [39] | 0.874 | 0.884 | 0.896 | 0.825 | 0.951 | 0.961 | 0.776 | 0.669 | 0.933 | 0.737 | 0.851 |
| FusAct10 [39] | 0.864 | 0.884 | 0.899 | 0.821 | 0.951 | 0.959 | 0.776 | 0.671 | 0.929 | 0.748 | 0.850 |
| SegNet [22] | 0.73 | 0.813 | 0.802 | 0.737 | 0.889 | 0.869 | 0.708 | 0.328 | - | - | - |
| U-Net [22] | 0.787 | 0.779 | 0.713 | 0.686 | 0.848 | 0.836 | 0.671 | 0.332 | - | - | - |
| DeepLabV3+ [22] | 0.875 | 0.879 | 0.899 | 0.817 | 0.939 | 0.954 | 0.774 | 0.628 | - | - | - |

The results shown in Table 10 confirm previous conclusions drawn in the polyp experiments. HN&101 is our suggested approach.

## 4 CONCLUSIONS

Semantic segmentation became crucial for medical-image analysis. Indeed, being able to identify anomalies in the human body and circumscribe their boundaries is fundamental for beneficial therapies, especially in the early stage of a disease.

In this work, we propose the use of an ensemble of networks for the purpose of semantic segmentation. The ensemble is built by varying the loss functions of the internal networks, this is done to enforce diversity among the individual segmentators.

We have tested our approach with several loss functions and we evaluate the framework on two real-world scenarios, namely in the polyp and skin segmentation areas.

We provide an in-depth empirical analysis demonstrating that our approach results in good and promising performance. The empirical evaluation comprises also a comparison with the most recent state of the art approaches, e.g. HarDNet-MSEG [24] and transformer-based approaches.

As a future work, we plan to reduce the complexity of the ensemble. This can be reach applying different techniques, for instance: pruning, quantization, low-rank factorization and distillation.

The code of the described experiments as well as all the descriptors reported in this paper is publicly available at https://github.com/LorisNanni.

## ACKNOWLEDGEMENT

Through their GPU Grant Program, NVIDIA donated the TitanX GPU that was used to train the CNNs presented in this work.

**Loris Nanni** is an associate professor at the Department of Information Engineering of the University of Padova; he is in charge of the "Operating Systems" and "Artificial Intelligence" courses at the degree program in Computer Science.
His research activity focuses on pattern recognition and machine learning, in particular he mainly works in biometrics systems. He has an H-index of 52 (google scholar)
Loris Nanni received his Master degree cum laude in the June 2002, his PhD in computer engineering in the April 2006. He is an associate researcher since March 2011 and he is an associate professor since March 2017.

**Alessandra Lumini** received her Master's Degree cum laude from the University of Bologna in 1996, and in 2001 she received her Ph.D. degree in Computer Science for her work on "Image Databases". She is currently an Associate Professor at DISI, University of Bologna. She is a member of the Biometric Systems Lab and of the Smart City Lab and she is interested in biometric systems, pattern recognition, machine learning, image databases, and bioinformatics. She is coauthor of more than 200 research papers. Her Google H-index is 45.

**Daniela Cuza** received her Bachelor Degree in Biomedical Engineering from the University of Padua in 2021. She is currently a Master Degree student in Computer Engineering at the University of Padua. Her research interests mainly focus on deep learning and machine learning.

**Andrea Loreggia** is an Assistant Professor at the University of Bologna. His research interests in artificial intelligence span from knowledge representation to deep learning. His studies are dedicated to designing and providing tools for the development of intelligent agents capable of representing and reasoning with preference and ethical-moral principles. Member of the UN/CEFACT group of experts, he actively participates in the dissemination and sustainable development of technology. He received his Master's Degree cum laude from the University of Padova in 2012, and in 2016 he received his Ph.D. degree in Computer Science.

**Sheryl B. Brahnam** is professor of computer information systems. Her research interests include decision support systems, artificial intelligence and computer vision, modeling and simulation, cultural and ethical aspects of technology, and rhetoric and conversational agents. Her teaching interests are in the areas of management information systems, decision support systems, distance education, programming languages and Internet for business. She has more than 140 publications in academic journals, book chapters and conference proceedings.